\documentclass{article}
\usepackage[final]{colm2026_conference}

\usepackage{lineno}

\usepackage[utf8]{inputenc}
\usepackage[T1]{fontenc}
\usepackage{hyperref}
\usepackage{url}
\usepackage{booktabs}
\usepackage{amsfonts}
\usepackage{amsmath}
\usepackage{amssymb}
\usepackage{nicefrac}
\usepackage{microtype}
\usepackage{graphicx}
\usepackage{xcolor}
\usepackage{listings}
\usepackage{multirow}
\usepackage{enumitem}
\usepackage{float}
\usepackage{subcaption}
\usepackage{pifont}
\usepackage{colortbl}
\usepackage{mdframed} 
\usepackage{tcolorbox}
\tcbuselibrary{skins, breakable}

\definecolor{darkblue}{rgb}{0, 0, 0.5}
\hypersetup{colorlinks=true, citecolor=darkblue, linkcolor=darkblue, urlcolor=darkblue}

\definecolor{darkgreen}{RGB}{0,128,0}
\definecolor{darkred}{RGB}{180,0,0}

\lstdefinelanguage{pylang}{
  keywords={function, return, if, else, while, print},
  keywordstyle=\color{blue}\bfseries,
  ndkeywords={len},
  ndkeywordstyle=\color{teal}\bfseries,
  sensitive=true,
  comment=[l]{//},
  morecomment=[s]{/*}{*/},
  commentstyle=\color{gray}\ttfamily,
  stringstyle=\color{red}\ttfamily,
  morestring=[b]",
}

\title{Syntax Without Semantics: Teaching Large Language Models to Code in an Unseen Language}

\author{%
  Vinayshekhar Bannihatti Kumar\textsuperscript{*}, Disha Makhija\textsuperscript{*}, Manoj Ghuhan Arivazhagan \\
  \textbf{Rashmi Gangadharaiah} \\
  AWS AI Labs \\
  \texttt{\{vinayshk,dismakhi,mghuhan,rgangad\}@amazon.com}
}

\begin{document}

\ifcolmsubmission
\linenumbers
\fi

\maketitle
\renewcommand{\thefootnote}{\fnsymbol{footnote}}
\footnotetext[1]{\textsuperscript{}Equal contribution.}
\renewcommand{\thefootnote}{\arabic{footnote}}

\begin{abstract}
Large language models (LLMs) achieve high pass rates on code generation benchmarks, yet whether they can transfer this ability to languages absent from pretraining remains poorly understood. We introduce PyLang, a minimal imperative language absent from all pretraining corpora, and evaluate frontier models zero-shot and fine-tuned Qwen3 (4B, 8B, 32B) on 352 problems. We find that fine-tuning quickly teaches syntax but fails to transfer semantic competence: Python outperforms PyLang by up to 19\% across all configurations, and no intervention (multi-task learning, preference tuning, code infilling, or latent-space objectives) closes the gap. An LLM judge reveals that frontier models select an identical algorithm to Python 80\% of the time, yet cannot translate it into a working PyLang implementation, and CKA analysis confirms that fine-tuned models converge to nearly identical internal representations across languages (CKA~$>$~0.97) while diverging at the output stage. We term this the \emph{implementation fidelity} gap: models possess language-agnostic algorithmic understanding but cannot express it in an unfamiliar language. Our findings highlight the need for training methods that decouple reasoning from language-specific realization.

\end{abstract}

\begin{center}
    \small \url{https://github.com/amazon-science/pylang}
\end{center}

\section{Introduction}
\label{sec:intro}

Large language models (LLMs) achieve high pass rates on code generation benchmarks and power production coding assistants used by millions of developers~\citep{chen2021evaluating, roziere2023code, li2023starcoder}. Yet real-world demand extends well beyond pretraining corpora: domain-specific languages, proprietary syntaxes, and entirely new languages such as Mojo and Carbon continue to emerge. Cross-lingual evaluations have shown that model performance drops sharply on low-resource languages~\citep{cassano2023multipl, baltaji2023crosslingual}, and recent work on esoteric languages suggests that LLM reasoning may overfit to training syntax~\citep{sharma2026esolangbench, han2025reward}. For LLMs to be useful in these settings, they must transfer the algorithmic reasoning acquired during pretraining to new languages, not merely learn their surface syntax. If these skills remain locked to familiar languages, the practical value of LLMs for new and emerging languages is severely limited.

This expectation rests on an implicit assumption: that pretraining endows models with \emph{general algorithmic reasoning} separable from any particular language, which can be composed with newly learned syntax to produce correct code in the target language. If the assumption holds, adaptation is straightforward: teach the model a new grammar and a handful of idioms, and existing algorithmic knowledge transfers. If it does not, and code generation ability is deeply entangled with the specific languages seen during pretraining, then every new language demands far more data and compute than practitioners expect, and the benefits of pretraining remain confined to the languages it already covers.

Distinguishing between these two possibilities requires answering a concrete question: \emph{When we fine-tune a model on code in a new language, what does it actually learn?} Does it acquire genuine problem solving ability in the new language, or does it merely learn the syntax, leaving its algorithmic knowledge bound to familiar languages? Answering this question with existing languages is difficult: fine-tuning on Python or Java confounds new learning with the billions of lines of the same language already seen during pretraining~\citep{cassano2023multiplt, giagnorio2025nosilverbullet}.

We resolve this confound by introducing \textbf{PyLang}, a purpose-built programming language unseen in any pretraining corpus. PyLang uses familiar C/JavaScript syntax but strips away \texttt{for} loops, the standard library, and all high-level abstractions, so that learning the grammar is easy but writing correct solutions requires building common operations from scratch (\S\ref{sec:pylang}). We fine-tune separate Qwen~3 models~\citep{yang2025qwen3technicalreport} on Python and PyLang solutions to the same problems. The Python model's ability reflects pretraining \emph{plus} fine-tuning; the PyLang model's ability derives \emph{entirely} from fine-tuning. Comparing the two cleanly isolates how much algorithmic reasoning actually transfers to a new language.

Our contributions are:

\begin{enumerate}[leftmargin=*,nosep]
  \item \textbf{PyLang: a controlled testbed for cross-language transfer.} We design and release PyLang, a minimal Turing-complete language absent from all pretraining corpora, along with parallel PyLang and Python solutions for 2{,}250 training problems and 352 test problems (251 in-distribution Codeforces~\citep{penedo2025codeforces} and 101 out-of-distribution MBPP~\citep{austin2021program}). PyLang's familiar syntax but deliberately impoverished semantics separate what models can learn from fine-tuning alone versus what they inherit from pretraining, providing a reusable benchmark for future research on code generation (\S\ref{sec:pylang}).

 \item \textbf{Systematic evaluation across scales, methods, and model families.} We evaluate frontier models (Sonnet~4.5, Sonnet~4, GPT~5.4) zero-shot and fine-tune open-source Qwen~3 models (4B, 8B, 32B) under two SFT strategies (direct question-to-code and multi-task with intermediate reasoning~\citep{jiang2023selfplanning}), two preference tuning methods (DPO~\citep{rafailov2023direct} and GRPO~\citep{shao2024deepseekmath}), and two targeted interventions (code infilling and JEPA). The gap exists even in frontier models but is far worse in open-source models: Sonnet~4.5 shows a 30\% deficit while fine-tuned Qwen models show 11-34\% gaps across all configurations (\S\ref{sec:results}, \S\ref{sec:interventions}).

  \item \textbf{Diagnosis of the bottleneck through converging evidence.} We identify \emph{implementation fidelity}, the ability to realize a correct algorithm as working code in a given language, as the primary bottleneck, supported by three lines of evidence: (i)~an LLM judge shows that frontier models select the same algorithm in PyLang and Python 80\% of the time, yet the PyLang implementation fails to pass test cases; (ii)~error decomposition reveals that 75\% of failures involve the correct algorithm with buggy implementation, not a wrong algorithmic choice; and (iii)~mechanistic interpretability confirms that fine-tuned PyLang and Python models converge to nearly identical internal representations (CKA~$>$~0.97) yet diverge at the output stage (\S\ref{sec:judge}, \S\ref{sec:interp}).
\end{enumerate}

Together, these contributions establish that LLMs possess language-agnostic algorithmic understanding but cannot express it in unfamiliar languages~\citep{basu2026interpretability}, and provide a foundation and benchmark for developing training methods that decouple reasoning from language-specific realization.

\section{PyLang: Language Design and Rationale}
\label{sec:pylang}

Isolating the contribution of fine-tuning from that of pretraining requires a target language satisfying two properties: (i)~it must be \emph{absent} from all pretraining corpora, so that any capability in the language derives entirely from fine-tuning, and (ii)~it must be \emph{Turing complete}, so that every problem solvable in Python is also solvable in the target language and cross-language comparison is meaningful.
We designed PyLang to satisfy both requirements while creating a controlled testbed that separates syntax acquisition from semantic competence. PyLang is a minimal imperative language whose design reflects three deliberate choices.

\textbf{Familiar surface syntax:} PyLang borrows curly brace block delimiters, semicolon terminated statements, and the \texttt{function} keyword from the C/JavaScript family. These elements are well represented in pretraining corpora and allow the model to leverage existing tokenization and syntactic pattern knowledge. The intent is to make syntax acquisition \emph{easy}: if the model still fails after learning the grammar, the failure must lie elsewhere.

\textbf{Deliberately impoverished semantics.}
PyLang omits \texttt{for} loops, list comprehensions, classes, exception handling, and the entire standard library. All iteration must use \texttt{while}; all string to integer conversion, tokenization, and sorting must be implemented manually from primitive operations. This creates a large gap between syntactic competence (writing code that parses) and semantic competence (writing code that computes the correct answer), which is precisely the gap we wish to measure.

\textbf{Controlled feature set:} Table~\ref{tab:pylang_features} enumerates the supported and excluded features, along with an illustrative example showing the practical consequence. The included primitives (integer arithmetic, string concatenation, \texttt{while} loops, conditionals, user defined functions, dictionary based arrays, \texttt{print()}, and \texttt{len()}) makes PyLang Turing complete. The excluded features represent the high level abstractions that make Python concise and that models have extensive experience composing. By removing them, we force the model to reconstruct common operations from first principles, testing whether algorithmic knowledge acquired during pretraining can be expressed through unfamiliar low level constructs.
The pair counting example illustrates this gap: the algorithmic core is identical in both languages, but PyLang's lack of \texttt{int()}, \texttt{split()}, and \texttt{for} loops increases the surface area for implementation errors without changing the underlying algorithmic challenge.

\begin{table}[h]
\centering
\caption{Feature comparison and illustrative example showing PyLang's (PL) deliberately impoverished semantics.}
\label{tab:pylang_features}
\scriptsize
\begin{minipage}[t]{0.28\textwidth}
\vspace{0pt}
\begin{tabular}{lcc}
\toprule
\textbf{Feature} & \textbf{PL} & \textbf{Py} \\
\midrule
Arithmetic & \checkmark & \checkmark \\
Strings & \checkmark & \checkmark \\
Variables & \checkmark & \checkmark \\
\texttt{while} & \checkmark & \checkmark \\
\texttt{if/else} & \checkmark & \checkmark \\
Functions & \checkmark & \checkmark \\
Dicts & \checkmark & \checkmark \\
\texttt{print/len} & \checkmark & \checkmark \\
\midrule
\texttt{for} & \ding{55} & \checkmark \\
Comprehensions & \ding{55} & \checkmark \\
Classes & \ding{55} & \checkmark \\
Exceptions & \ding{55} & \checkmark \\
Stdlib & \ding{55} & \checkmark \\
\bottomrule
\end{tabular}
\end{minipage}
\hfill
\begin{minipage}[t]{0.34\textwidth}
\vspace{0pt}
\begin{lstlisting}[language=Python, title=Python, basicstyle=\fontsize{5.5}{6.5}\ttfamily, xleftmargin=0pt, aboveskip=2pt, belowskip=2pt, numbers=none]
def solve(a, n):
  total = sum(a)
  count = 0
  for i in range(n):
    for j in range(i+1,n):
      if n*(a[i]+a[j]) == 2*total:
        count += 1
  print(count)
\end{lstlisting}
\end{minipage}
\hfill
\begin{minipage}[t]{0.34\textwidth}
\vspace{0pt}
\begin{lstlisting}[language=pylang, title=PyLang, basicstyle=\fontsize{5.5}{6.5}\ttfamily, xleftmargin=0pt, aboveskip=2pt, belowskip=2pt, numbers=none]
function solve(a, n) {
  count = 0; i = 0; sum = 0; t = 0;
  while (t < n) {
    sum = sum + a[t];
    t = t + 1;
  }
  while (i < n) {
    j = i + 1;
    while (j < n) {
      if (n*(a[i]+a[j]) == 2*sum) {
        count=count+1;
      }
      j = j + 1;
    }
    i = i + 1;
  }
  print(count);
}
\end{lstlisting}
\end{minipage}
\end{table}

The PyLang interpreter can be organized into three stages: a lexer that tokenizes the input into a stream of typed tokens, a recursive descent parser that produces an abstract syntax tree (AST), and a tree walking evaluator that executes the AST under lexical scoping rules.
The interpreter supports first class functions, nested scopes, and mutable dictionary based arrays. See Appendix~\ref{app:pylang} for the full PyLang specification and additional examples.

\section{Experimental Design and Results}
\label{sec:experimental}

\subsection{Benchmark, Models and Training}
We evaluate on 352 problems from two sources: Codeforces (251 problems; competitive programming tasks overlapping with the fine-tuning distribution, serving as our in-distribution benchmark) and MBPP (101 problems; utility tasks held entirely out of training, serving as our out-of-distribution benchmark). Each data-point includes a problem language statement, a verified reference solution, and a test suite (771 test cases total).

We use decoder-only models from the \textbf{Qwen~3} family~\citep{yang2025qwen3technicalreport} at three scales: \textbf{4B}, \textbf{8B}, and \textbf{32B} parameters to test across different model scales. We first evaluate all three models zero-shot, then fine-tune them using an identical training configuration: AdamW ($\text{lr} = 2 \times 10^{-5}$, weight decay $0.01$), 3 epochs, BF16 mixed precision, DeepSpeed ZeRO Stage~3 on 8$\times$A100 GPUs, sequence length 4{,}096, gradient clipping at 1.0. We train for 3 epochs, evaluating the checkpoint at the end of each epoch, and report results from the best-performing checkpoint. We also evaluate GPT 5.4, Claude Sonnet~4 and Sonnet~4.5 zero-shot via API, prompted with different prompt variations to measure how well state-of-the-art models handle an unseen language without parameter updates, and provide an anchor for interpreting the fine-tuned models' cross-language gap.

\subsection{Training Data and Strategies}

The fine-tuning corpus consists of 2{,}250 Codeforces-style problems, each with verified solutions in both PyLang and Python. All 101 MBPP problems are held out for out-of-distribution evaluation. For each strategy below, we train \emph{separate} PyLang and Python models on identically structured data from the same source problems. Appendix~\ref{app:datagen} details the data-generation pipeline. 

\textbf{Strategy~1: (Q$\rightarrow$C) } Each training example directly maps a natural language problem specification to executable code (2{,}250 per language). This is our minimal baseline investigating whether a simple input-output mapping, without intermediate reasoning steps, provides sufficient signal for models to acquire both language syntax and solution semantics.

\textbf{Strategy~2: Multi-Task Learning with Intermediate Reasoning (Q$\to$S, S$\to$C, Q$\to$C).}  Following principles from chain-of-thought prompting~\citep{wang2024plansearch,jiang2023selfplanning}, each problem generates three training instances per language (6{,}750 total): question-to-solution, solution-to-code, and question-to-code. The solution here refers to the natural language description of the correct algorithm. The intermediate reasoning step encourages the model to articulate an algorithmic strategy before generating code.

\textbf{Strategy~3: Preference Tuning (DPO and GRPO).}
The previous strategies train the model to imitate reference solutions. An alternative hypothesis is that the model \emph{can} produce correct code but assigns insufficient probability to it.
We explore two preference tuning methods, both initialized from the Q$\to$C SFT checkpoint at 4B and 8B scales:
DPO~\citep{rafailov2023direct} trains on $4{,}400$ offline preference pairs (passing vs.\ failing completions sampled from the SFT model), while
GRPO~\citep{shao2024deepseekmath} performs online exploration with execution-based rewards ($r = \text{tests\_passed} / \text{tests\_total}$).
We use a fractional reward rather than binary pass/fail to provide a denser training signal, particularly for PyLang where fully correct solutions are rare and partial credit helps the model learn incrementally. Details are in Appendix~\ref{app:hyperparams}.

\subsection{Results}
\label{sec:results}

\begin{table*}[t]
\centering
\caption{Performance comparison (in \%) across PyLang and Python on Codeforces and MBPP benchmarks (best checkpoint across 3 epochs).}
\label{tab:results}
\resizebox{\textwidth}{!}{
\begin{tabular}{l|cc|cc|cc|cc|cc|cc}
\toprule
& \multicolumn{4}{c|}{\textbf{Overall}} & \multicolumn{4}{c|}{\textbf{Codeforces}} & \multicolumn{4}{c}{\textbf{MBPP}} \\
\cmidrule(lr){2-5} \cmidrule(lr){6-9} \cmidrule(lr){10-13}
& \multicolumn{2}{c|}{\textbf{PyLang}} & \multicolumn{2}{c|}{\textbf{Python}} & \multicolumn{2}{c|}{\textbf{PyLang}} & \multicolumn{2}{c|}{\textbf{Python}} & \multicolumn{2}{c|}{\textbf{PyLang}} & \multicolumn{2}{c}{\textbf{Python}} \\
\cmidrule(lr){2-3} \cmidrule(lr){4-5} \cmidrule(lr){6-7} \cmidrule(lr){8-9} \cmidrule(lr){10-11} \cmidrule(lr){12-13}
\textbf{Model} & \textbf{Prob.} & \textbf{Syn.} & \textbf{Prob.} & \textbf{Syn.} & \textbf{Prob.} & \textbf{Test} & \textbf{Prob.} & \textbf{Test} & \textbf{Prob.} & \textbf{Test} & \textbf{Prob.} & \textbf{Test} \\
& \textbf{Pass} & \textbf{Err} & \textbf{Pass} & \textbf{Err} & \textbf{Pass} & \textbf{Pass} & \textbf{Pass} & \textbf{Pass} & \textbf{Pass} & \textbf{Pass} & \textbf{Pass} & \textbf{Pass} \\
\midrule
\multicolumn{13}{l}{\textit{Frontier Models}} \\
\midrule
Sonnet 4 (w/ snippets)       & 46.6 & 27.0 & 85.5 & 0.0 & 41.8 & 38.5 & 86.5 & 88.9 & 58.4 & 67.0 & 83.2 & 85.5 \\
Sonnet 4.5 (w/ snippets)     & 58.0 & 9.0 & 87.5 & 0.0 & 55.4 & 60.7 & 88.8 & 91.9 & 64.4 & 73.3 & 84.2 & 85.8 \\
GPT 5.4 (w/ snippets)     & 51.1 & 10 & 78.7 & 0.0 & 43.4 &  74.9 & 45.5 & 81.6 & 70.3 & 74.9 & 83.1 & 85.1\\
\midrule
\multicolumn{13}{l}{\textit{Base Models (no finetuning)}} \\
\midrule
Qwen 4B (base)                & 0.0 & 100 & 47.7 & 4.2 & 0.0 & 0.0 & 40.2 & 50.2 & 0.0 & 0.0 & 66.3 & 69.3 \\
Qwen 8B (base)                & 0.0 & 94.6 & 52.8 & 2.0 & 0.0 & 0.0 & 45.0 & 55.8 & 0.0 & 0.0 & 72.3 & 76.9 \\
Qwen 32B (base)               & 7.4 & 77.2 & 54.0 & 0.3 & 0.4 & 0.2 & 44.6 & 55.8 & 24.8 & 26.4 & 77.2 & 81.8 \\
\midrule
\multicolumn{13}{l}{\textit{Question $\rightarrow$ Code}} \\
\midrule
Qwen 4B                      & 36.1 & 4.5 & 47.7 & 3.7 & 32.7 & 44.2 & 40.2 & 49.8 & 44.5 & 53.1 & 66.3 & 70.0 \\
Qwen 8B                      & 38.6 & 4.2 & 57.4 & 0.0 & 32.7 & 42.3 & 53.0 & 60.3 & 53.5 & 64.4 & 68.3 & 72.3 \\
Qwen 32B                     & 47.2 & 2.5 & 63.6 & 0.3 & 48.2 & 57.7 & 60.2 & 66.2 & 44.5 & 56.8 & 72.3 & 75.2 \\
\midrule
\multicolumn{13}{l}{\textit{Q $\rightarrow$ S, S $\rightarrow$ C, Q $\rightarrow$ C (Combined)}} \\
\midrule
Qwen 4B                      & 35.5 & 3.4 & 44.3 & 0.0 & 29.9 & 42.5 & 37.9 & 46.8 & 49.5 & 58.1 & 60.4 & 64.0 \\
Qwen 8B                      & 38.9 & 3.7 & 43.8 & 0.3 & 33.1 & 44.9 & 40.6 & 47.9 & 53.5 & 64.4 & 51.5 & 57.1 \\
Qwen 32B                     & 50.6 & 1.1 & 52.0 & 9.1 & 47.8 & 58.3 & 47.4 & 56.4 & 57.4 & 65.0 & 63.4 & 67.3 \\
\midrule
\multicolumn{13}{l}{\textit{GRPO (Question $\rightarrow$ Code)}} \\
\midrule
Qwen 4B       & 37.8 & 4.0 & 54.0 & 3.1 & 34.3 & 43.6 & 47.0 & 58.3 & 46.5 & 58.1 & 71.3 & 74.9 \\
Qwen 8B     & 41.5 & 2.6 & 60.2 & 0.0 & 36.7 & 48.5 & 57.4 & 66.0 & 53.5 & 63.4 & 67.3 & 70.6 \\
\midrule
\multicolumn{13}{l}{\textit{DPO (Question $\rightarrow$ Code)}} \\
\midrule
Qwen 4B       & 35.8 & 4.0 & 55.1 & 0.9 & 31.5 & 42.9 & 50.6 & 61.5 & 46.5 & 53.1 & 66.3 & 68.3 \\
Qwen 8B       & 39.8 & 3.4 & 61.9 & 0.0 & 34.3 & 45.7 & 58.2 & 67.5 & 53.5 & 63.0 & 71.3 & 75.9 \\
\midrule
\multicolumn{13}{l}{\textit{Infilling + Q $\rightarrow$ C}} \\
\midrule
Qwen 4B                      & 35.8 & 3.1 & 45.2 & 4.2 & 31.5 & 41.5 & 38.6 & 48.5 & 46.5 & 57.4 & 61.4 & 65.3 \\
Qwen 8B                      & 38.4 & 4.8 & 49.4 & 0.28 & 33.5 & 45.3 & 42.6 & 51.5 & 50.5 & 60.4 & 66.3 & 70.0 \\
Qwen 32B                     & 49.4 & 1.14 & 64.2 & 0.6 & 47.8 & 58.3 & 58.2 & 65.6 & 53.5 & 63.7 & 79.2 & 82.2 \\
\midrule
\multicolumn{13}{l}{\textit{JEPA-based Training (Combined Dataset)}} \\
\midrule
Qwen 4B                      & 33.5 & 7.4 & 43.8 & 0.6 & 30.3 & 37.8 & 37.9 & 41.5 & 41.6 & 50.2 & 58.4 & 62.0 \\
Qwen 8B                      & 39.5 & 1.4 & 48.3 & 2.5 & 34.7 & 45.9 & 44.2 & 54.3 & 51.5 & 60.7 & 58.4 & 62.0 \\
\bottomrule
\end{tabular}
}
\end{table*}


We evaluate all generated solutions against the test suites using the respective language runtimes (PyLang interpreter or Python~3.11) and report three metrics: \textbf{Problem Pass Rate} (percentage of problems where all tests pass), \textbf{Test Pass Rate} (percentage of individual test cases passed capturing partial correctness), and \textbf{Syntax Errors} (percentage of problems that fail to parse indicating incomplete grammar acquisition). For evaluating zero-shot, we include the full interpreter specification in the prompt, providing the model with complete knowledge of PyLang’s syntax and semantics.  Table~\ref{tab:results} summarizes our main results.

\textbf{Without fine-tuning, the gap is stark.} Base Qwen3 models score 0-7\% on PyLang while reaching 48-54\% in Python, producing hundreds of syntax errors per run. Even frontier models given the full language specification \emph{and} starter code for input parsing fall well short with Sonnet~4.5 scoring 58\% in PyLang versus 88\% in Python, a 30\% deficit. Neither more demonstrations nor more reasoning closes it. Three-shot in-context prompting leaves Qwen3-32B at just 10.2\% on PyLang (Table~\ref{tab:icl}), far below its fine-tuned 47.2\%, showing that a handful of exemplars cannot override the Python pretraining prior. Nor is the gap an artifact of inference mode, with extended thinking enabled, Sonnet~4.5's gap slightly widens (29.5$\to$30.4~pp) and Sonnet~4's is essentially unchanged (38.9$\to$38.3~pp, Table~\ref{tab:thinking}). Extra reasoning tokens refine \emph{what} to do, but the failure lies in \emph{how} to express it in an unfamiliar language, so the added compute is absorbed by the Python setting, where the implementation path is already fluent, widening the cross-language gap rather than closing it.

\begin{table}[t]
\centering
\caption{Extended thinking vs.\ non-thinking for frontier models (352 problems, snippet prompt of Table~\ref{tab:results}). Thinking disproportionately helps Python, leaving the cross-language gap intact.}
\label{tab:thinking}
\small
\begin{tabular}{llccc}
\toprule
\textbf{Model} & \textbf{Mode} & \textbf{PyLang} & \textbf{Python} & \textbf{Gap} \\
\midrule
Sonnet 4   & no-think & 46.6 & 85.5 & 38.9 \\
Sonnet 4   & think    & 50.9 & 89.2 & 38.3 \\
Sonnet 4.5 & no-think & 58.0 & 87.5 & 29.5 \\
Sonnet 4.5 & think    & 61.4 & 91.8 & 30.4 \\
\bottomrule
\end{tabular}
\end{table}

\begin{table}[t]
\centering
\caption{Three-shot in-context PyLang prompting on \emph{base} Qwen3 (352 problems), with 3 verified PyLang solutions as exemplars. All results fall far below the fine-tuned pass rates (36.1/38.6/47.2\% for 4B/8B/32B, Table~\ref{tab:results}); the high syntax-error counts show the model largely fails to produce parseable PyLang.}
\label{tab:icl}
\small
\begin{tabular}{lccc}
\toprule
\textbf{Model} & \textbf{Prob.\ Pass \%} & \textbf{Test Pass \%} & \textbf{Syn.\ Err} \\
\midrule
Qwen3-4B + 3-shot  & 1.1 (4/352)   & 1.3  & 759 \\
Qwen3-8B + 3-shot  & 2.6 (9/352)   & 3.2  & 740 \\
Qwen3-32B + 3-shot & 10.2 (36/352) & 14.9 & 617 \\
\bottomrule
\end{tabular}
\end{table}

\textbf{Fine-tuning teaches syntax but does not close the gap.} Fine-tuning dramatically reduces syntax errors and lifts PyLang performance: in the Q~$\rightarrow$~C setting, Qwen~32B reaches 47.2\% in PyLang, up from 7.4\% at baseline. However, the gap with Python persists across every model size and training configuration. Under Q~$\rightarrow$~C, the gap ranges from 11.6\% (4B) to 18.8\% (8B), and scaling from 4B to 32B improves both languages but improves Python \emph{more} (+15.9\% vs.\ +11.1\%), so the gap does not shrink with scale. Training strategies shift the gap without eliminating it. The Combined objective appears to close the gap at 32B (50.6\% PyLang vs.\ 52.0\% Python), but this is because the multi-task objective hurts Python performance rather than helping PyLang. Across all other configurations, Python remains substantially ahead. On out-of-distribution MBPP problems the picture is starker: even the strongest configuration (Combined~32B) shows Python outperforming PyLang by 6\%, while Q~$\rightarrow$~C gaps reach 22-28\% at smaller scales. For fine-tuned models, syntax errors drop to less than 5\%, confirming that the residual gap is not syntactic and that models learn to write code that \emph{parses} in PyLang but not code that \emph{works}, i.e., passes test cases.

\textbf{Preference tuning does not close the gap.} We further investigate whether post-SFT alignment can bridge the correctness gap by applying two preference optimization methods, DPO and GRPO, on top of the Q$\to$C SFT checkpoints at 4B and 8B scales. Both methods improve Python more than PyLang. DPO~8B achieves the highest Python result at 61.9\% (+4.5\ \% over the SFT baseline), while GRPO~8B reaches 60.2\% (+2.8\ \%). PyLang gains are more modest: GRPO~8B reaches 41.5\% (+2.9\ \%) and DPO~8B reaches 39.8\% (+1.2\ \%). The gap \emph{widens} from 18.8 \% (SFT) to 22.1 \% (DPO) at the 8B scale, and the pattern is consistent across both model sizes. The asymmetry arises because preference tuning can only upweight correct solutions that already exist in the model's sample distribution. Python's SFT checkpoint produces correct solutions frequently, providing rich training signal; PyLang's checkpoint produces them far less often, limiting what either DPO or GRPO can learn from.

\textbf{More data does not close the gap.} To test whether the residual gap simply reflects too little PyLang data, we sweep the SFT set from 10\% to 100\% of the 2{,}250 examples for 4B and 8B models(Table~\ref{tab:datascale}). PyLang performance saturates well before the full SFT budget: from 50\% to 100\% the 4B model gains only 4~pp (30.4\%$\to$34.7\%) and the 8B model under 2~pp (36.6\%$\to$38.4\%), leaving an asymptote 13--20~pp below the Python pass rate (47.7\% at 4B, 57.4\% at 8B). This is the regime practitioners face for genuinely new languages, where large paired corpora do not exist, and it is exactly what the implementation-fidelity hypothesis predicts: the bottleneck is not how many examples the model has seen but its inability to translate pretrained algorithmic knowledge into PyLang.  

\begin{table}[t]
\centering
\caption{PyLang problem pass rate (\%) vs.\ fraction of the 2{,}250-example training set. Performance saturates far below Python (47.7\% for 4B, 57.4\% for 8B).}
\label{tab:datascale}
\small
\begin{tabular}{lcccccc}
\toprule
\textbf{Training data \%} & \textbf{10} & \textbf{25} & \textbf{50} & \textbf{75} & \textbf{90} & \textbf{100} \\
\midrule
Qwen3-4B & 19.9 & 26.1 & 30.4 & 33.1 & 32.8 & 34.7 \\
Qwen3-8B & 28.1 & 33.0 & 36.6 & 37.4 & 37.7 & 38.4 \\
\bottomrule
\end{tabular}
\end{table}

\section{Understanding the Gap}
\label{sec:analysis}

The main results establish that fine-tuning eliminates most of the syntax errors, yet a substantial correctness gap remains. We now investigate the \emph{nature} of this gap, whether it reflects a deficit in algorithmic reasoning or in implementation, and \emph{where} it concentrates across problem types.

\subsection{Is the Gap Algorithmic or Implementational?}
\label{sec:judge}
Having established that syntax errors are not the bottleneck, we ask a more pointed question: when a model fails in PyLang, does it fail because it chose the wrong algorithm, or because it chose the right algorithm but could not implement it? We use an LLM judge (based on Claude Opus~4.6) to compare each model's PyLang and Python solutions for the same problem and determine whether both use the same underlying algorithm. We validated the LLM judge on 50 human-annotated problem pairs (three annotators, Cohen's $\kappa$ = 0.81), finding 94\% agreement with the human majority label. Table~\ref{tab:judge} summarizes the results.

\begin{table*}[t]
\centering
\begin{minipage}[t]{0.52\textwidth}
\centering
\captionsetup{font=scriptsize}
\caption{Cross-language algorithmic agreement (LLM judge, Claude Opus~4.6). \textit{Same~\%} = pairs using the same algorithm. Frontier comparisons use Python as reference; fine-tuned comparisons are PyLang vs.\ Python.}
\label{tab:judge}
\scriptsize
\begin{tabular}{llccc}
\toprule
& \textbf{Comparison} & \textbf{Same} & \textbf{Diff} & \textbf{Same \%} \\
\midrule
\multicolumn{5}{l}{\textit{Sonnet 4.5 (cross-language)}} \\
\midrule
& Python vs Java             & 267 & 66 & 80.2\% \\
& Python vs C++              & 266 & 67 & 79.9\% \\
& Python vs PyLang (w/ snippets)      & 257 & 70 & 78.6\% \\
& Python vs PyLang (w/o snippets)   & 231 & 77 & 75.0\% \\
\midrule
\multicolumn{5}{l}{\textit{Fine-tuned Qwen Q$\rightarrow$C}} \\
\midrule
& Qwen 32B & 158 & 153 & 50.8\% \\
& Qwen 8B  & 116 & 190 & 37.9\% \\
& Qwen 4B  & 106 & 194 & 35.3\% \\
\bottomrule
\end{tabular}
\end{minipage}%
\hfill
\begin{minipage}[t]{0.44\textwidth}
\centering
\captionsetup{font=scriptsize}
\caption{Prompt ablation for Sonnet~4.5 on PyLang. Each row adds one component to the base prompt. All gains come from implementation patterns, not algorithmic guidance.}
\label{tab:ablation}
\scriptsize
\begin{tabular}{lc}
\toprule
\textbf{Prompt Configuration} & \textbf{Pass \%} \\
\midrule
Base (spec only)                  & 16.5 \\
+ Code snippets                   & 40.3 \\
+ Chained indexing rule            & 37.4 \\
+ Snippets + indexing rule         & 58.0 \\
\midrule
\multicolumn{2}{l}{\textit{Python baseline (same model)}} \\
\midrule
Sonnet 4.5 Python & 87.5 \\
\bottomrule
\end{tabular}
\end{minipage}
\end{table*}

\textbf{Explicit constraints are insufficient; implementation patterns are essential.} Despite receiving the complete PyLang specification through the interpreter in the prompt and an explicit list of constraints, frontier models routinely violate them and generate \texttt{for} loops the interpreter cannot parse, chained indexing (multi-dimensional indexing) that produces runtime errors, and calls to built-in functions that do not exist. \emph{Showing} the model how to work within PyLang's constraints helps far more than \emph{telling} it what the constraints are. The results in (Table~\ref{tab:ablation}) show code snippets demonstrating correct idioms improve Sonnet~4.5 by +24\%, while an explicit rule about
unsupported chained indexing adds a further +21\%. We report judge results and ablation results for Sonnet~4.5 (and Sonnet~4 in Appendix~\ref{app:sonnet4}); both exhibit the same qualitative pattern. The model knows what PyLang does not support but it cannot reliably avoid using it, perhaps because its generation patterns remain anchored to the syntax it encountered during pretraining and its inability to adapt to the newer patterns. Even after incorporating prompt components that address the identified error cases, Sonnet~4.5 reaches only 58\% in PyLang versus 87.5\% in Python.

\textbf{Frontier models maintain high algorithmic consistency across languages.} Sonnet~4.5 uses the same algorithm in Python and C++/Java 80\% of the time. When we replace C++/Java with PyLang, agreement drops only to 79\%, effectively matching the known-language baseline. For this evaluation, we provide the full PyLang interpreter specification in the prompt, while C++, Java, and Python are specified by name alone. Despite this asymmetry, the model reasons about PyLang in essentially the same way it reasons about languages it knows well. Yet despite choosing identical algorithm nearly four-fifths of the time, Sonnet~4.5 achieves only 16.5\% PyLang correctness in the base configuration. Adding code snippets and syntax rules lifts correctness to 58.0\% (+41.5\%) while \emph{increasing} algorithmic agreement by 3.6\%. In other words, the snippets did not change \emph{what} the model decided to do rather they changed whether it could \emph{do} it correctly. The directionality of the gap confirms this: 112 ($\sim 32\%$) problems fall in the Python-succeeds-PyLang-fails direction, but only 8 ($\sim 2\%$) in the reverse. The model almost never solves a problem in PyLang that it cannot also solve in Python. The reasoning is there; the implementation fluency is not.

\textbf{When frontier models fail in PyLang, the failure is primarily implementational.} Among problems where Sonnet~4.5 succeeds in Python but fails in PyLang, 77\% use the same algorithm in both languages. The model selected the correct approach but could not express it as working PyLang code. The prompt ablation (prompts in Appendix~\ref{app:prompts}) confirms this diagnosis: every component that improves performance supplies a language-specific implementation pattern, character-by-character input parsing, digit-by-digit conversion, flattened array indexing, rather than problem-level algorithmic guidance.

\textbf{Fine-tuned models agree when correct, but diverge when they fail.} Fine-tuned Qwen models show substantially lower algorithmic agreement than frontier models: 51\% for 32B, 38\% for 8B, and 35\% for 4B. This is not simply a consequence of lower PyLang correctness. Sonnet~4.5 without code snippets achieves only 16.5\% PyLang correctness which is less than half that of Qwen~8B (38.6\%)  yet maintains 75\% algorithmic agreement, nearly double
Qwen~8B's 38\%. The frontier model writes the same algorithm in broken syntax; the fine-tuned model writes different algorithms entirely. This divergence is not an artifact of separate training: the PyLang and Python training sets contain paired solutions to the same problems, and when both languages produce correct solutions, agreement rises to 73-86\% across all model sizes comparable to Sonnet's overall rate. The models \emph{can} converge on shared
approaches; \emph{they diverge specifically when implementation fails}. In contrast, frontier models maintain algorithmic consistency even when their implementations are incorrect. This suggests that the ability to reason about a problem independently of the ability to implement the solution is something pretraining at scale provides but fine-tuning alone does not fully recover.

\textbf{Error decomposition reveals concrete implementation barriers.} We further classify every problem in the PyLang-Python gap as either a \textit{implementation fidelity} (the model uses the same algorithm but fails to implement it correctly) or an \textit{algorithmic gap} (the model uses a different algorithm entirely). Table~\ref{tab:decomp} summarizes the results. For Sonnet~4.5, 75\% of failures are language barriers. The largest
category is same-algorithm wrong output (34\% of the gap), i.e., the model picks the right approach but produces buggy code such as incorrect input
parsing, off-by-one loop boundaries, or mishandled types. Chained array indexing accounts for a further 19\% (21 problems), and unsupported keywords and operators contribute 8\%. Only 16\% of Sonnet~4.5's failures involve a genuinely different algorithm; the remaining 9\% could not be confidently classified. Notably, Sonnet~4.5 handles chained indexing far better than Sonnet~4 (19\% vs.\ 42\% of the gap), yet the overall
gap barely shrinks because other implementation errors occur. The problem is not any single missing feature but a broad deficit in implementation fluency. Fine-tuned models show a different mix. Syntax errors are rare ($\leq$9 problems per model among gap problems specifically), confirming that fine-tuning teaches grammar effectively. The dominant failure mode is again same-algorithm wrong output, right algorithm and wrong code, accounting for 13-24\% of the gap across model sizes. The model knows what to do and still produces code that does not work.

\subsection{Where the Gap Concentrates}
\label{sec:where}

\begin{table*}[t]
\centering
\begin{minipage}[t]{0.48\textwidth}
\centering
\captionsetup{font=scriptsize}
\caption{PyLang--Python gap decomposition. Each gap problem (Python correct, PyLang incorrect) classified as \textit{language barrier} or \textit{algorithmic gap}.}
\label{tab:decomp}
\setlength{\tabcolsep}{3pt}
\scriptsize
\begin{tabular}{p{2cm}ccccc}
\toprule
& \textbf{Son.4} & \textbf{Son.4.5} & \textbf{32B} & \textbf{8B} & \textbf{4B} \\
\midrule
Gap size & 146 & 112 & 90 & 98 & 83 \\
\midrule
Language barrier   & 80\% & 75\% & 40\% & 27\% & 51\% \\
Algorithmic gap    & 12\%  & 16\% & 48\% & 58\% & 30\% \\
\midrule
\multicolumn{6}{l}{\textit{Language Barrier breakdown}} \\
\midrule
Chained indexing        & 42\% & 19\% & 1\%  & 0\%  & 0\%  \\
Wrong output            & 18\% & 34\% & 24\% & 13\% & 17\%\\
Unsupp. words   & 10\% & 2\%   & 1\%  & 0\%  & 1\%  \\
Other                   & 11\% & 21\% & 13\% & 13\% & 33\% \\
\bottomrule
\end{tabular}
\end{minipage}%
\hfill
\begin{minipage}[t]{0.48\textwidth}
\centering
\captionsetup{font=scriptsize}
\caption{PyLang--Python gap (in \%) by stdlib dependence. \textit{Low/Med/High} = increasing reliance on built-in functions absent from PyLang. \textit{Py} = Python pass rate.}
\label{tab:stdlib}
\setlength{\tabcolsep}{3pt}
\scriptsize
\begin{tabular}{lcccccc}
\toprule
& \multicolumn{2}{c}{\textbf{Low} (238)} & \multicolumn{2}{c}{\textbf{Med} (83)} & \multicolumn{2}{c}{\textbf{High} (31)} \\
\cmidrule(lr){2-3} \cmidrule(lr){4-5} \cmidrule(lr){6-7}
\textbf{Model} & \textbf{Py} & \textbf{Gap} & \textbf{Py} & \textbf{Gap} & \textbf{Py} & \textbf{Gap} \\
\midrule
Sonnet 4.5 & 87.4 & $-$28.2 & 90.4 & $-$34.9 & 80.6 & $-$25.8 \\
Qwen 32B   & 61.8 & $-$13.0 & 66.3 & $-$19.3 & 71.0 & $-$35.5 \\
Qwen 8B    & 54.6 & $-$17.2 & 57.8 & $-$15.7 & 77.4 & $-$38.7 \\
Qwen 4B    & 47.1 & $-$11.8 & 42.2 & $-$1.2  & 67.7 & $-$38.7 \\
\bottomrule
\end{tabular}
\end{minipage}
\end{table*}


\textbf{Standard library dependence amplifies the gap.} We use an LLM classifier to assign each problem a stdlib dependence level (low, medium, high) based on whether its reference Python solution requires built-in functions, such as \texttt{sorted()}, \texttt{split()}, or \texttt{int()}, that PyLang does not provide. On low-stdlib problems, where both languages have roughly equivalent expressive power, the fine-tuned model gap ranges from 12-17\%. On high-stdlib problems it widens to 36-39\% (Table~\ref{tab:stdlib}). Part of this increase reflects a genuine language expressiveness difference, PyLang lacks built-ins that Python provides. But the gap on low-stdlib problems cannot be attributed to missing features and represents a purer measure of implementation fidelity: even when PyLang and Python offer the same primitives, the model produces correct Python solutions far more often. The gap in Sonnet~4.5's performance remains greater than $25\%$ for all categories.

Appendix~\ref{app:length} reports additional control on output-length stratification ruling out text length as an alternative explanation. 


\subsection{Can Targeted Interventions Close the Gap?}
\label{sec:interventions}

The preceding analyses characterize the gap; we now ask whether targeted training interventions can close it. Each intervention tests a specific hypothesis about the bottleneck.

\textbf{Infilling.} If the bottleneck is structural understanding (block nesting, function composition), then code infilling~\citep{fried2023incoder} should help. Infilling does reduce syntax errors (2.5 \% to 1.1 \% for 32B) but does not improve functional correctness: 35.8\% vs. 36.1\% for 4B, 49.4\% vs. 47.2\% for 32B (Table~\ref{tab:results}). The model understands how PyLang programs are organized but makes errors in the low-level details of implementation.


\textbf{JEPA.} If the bottleneck is shallow internal representations, then the embedding-space prediction objective of LLM-JEPA~\citep{llmjepa} should help. It does not: the 4B model drops from 35.5\% to 33.5\%, and while 8B edges up slightly (38.9\% vs. 39.5\%), Python falls sharply from 57.4\% to 48.3\% (Table~\ref{tab:results}), shrinking the gap only by hurting the known language. This rules out representation quality as the bottleneck.


\medskip
\noindent
These results converge on a single explanation. The gap is not algorithmic (frontier models identify the correct algorithm but fail to implement it), not structural (infilling fixes syntax but not correctness), not representational. What remains is \emph{implementation fidelity}: the model understands what to do but cannot reliably produce the tokens that do it in an unfamiliar language.




\section{Interpretability Analysis}
\label{sec:interp}

To characterize \emph{how} fine-tuning reshapes internal representations, we compare three Qwen~3-8B checkpoints: the base (pretrained) model, PyLang-FT, and Python-FT.
Both fine-tuned models were trained for exactly 105 steps (3 epochs) on matched data, eliminating training duration as a confound.
All analyses use actual model generations on 60 held-out test problems with per-problem pass/fail labels. We score each model's generations under all three models (Table~\ref{tab:perplexity}) and measure Centered Kernel Alignment (CKA)~\citep{davari2022cka} between hidden representations at every layer (Table~\ref{tab:cka}).

\begin{table*}[t]
\centering
\begin{minipage}[t]{0.42\textwidth}
\centering
\captionsetup{font=scriptsize}
\caption{Cross-model perplexity on actual generations (lower $=$ more expected). Each row is a set of generations; each column is the evaluating model.}
\label{tab:perplexity}
\scriptsize
\begin{tabular}{lccc}
\toprule
& \multicolumn{3}{c}{\textbf{Model Evaluating}} \\
\cmidrule(lr){2-4}
\textbf{Generated by} & \textbf{Base} & \textbf{PyL-FT} & \textbf{Py-FT} \\
\midrule
PyLang-FT  & 1.50 & \textbf{1.10} & 1.44 \\
Python-FT  & 2.64 & 2.75 & \textbf{2.30} \\
\bottomrule
\end{tabular}
\end{minipage}%
\hfill
\begin{minipage}[t]{0.55\textwidth}
\centering
\captionsetup{font=scriptsize}
\caption{Layer-wise CKA similarity (1.0\,$=$\,identical). Both fine-tuned models diverge from base in later layers; the two fine-tuned models remain nearly identical.}
\label{tab:cka}
\scriptsize
\begin{tabular}{lccc}
\toprule
\textbf{Layer} & \textbf{Base vs PyL-FT} & \textbf{Base vs Py-FT} & \textbf{PyL-FT vs Py-FT} \\
\midrule
0  & 1.000 & 1.000 & 1.000 \\
12 & 0.987 & 0.985 & 0.998 \\
18 & 0.941 & 0.963 & 0.968 \\
24 & 0.939 & 0.957 & 0.981 \\
30 & 0.849 & 0.860 & 0.995 \\
35 & \textbf{0.631} & \textbf{0.661} & 0.990 \\
\bottomrule
\end{tabular}
\end{minipage}
\end{table*}

Cross-model perplexity reveals that PyLang-FT reduces perplexity on its own generations by 27\% (1.50\,$\to$\,1.10), whereas Python-FT improves by only 13\% (2.64\,$\to$\,2.30), confirming that PyLang fine-tuning restructures the model far more aggressively than Python fine-tuning.

CKA analysis tells the complementary story. At layer~35, PyLang-FT retains only 63.1\% of the base model's representational structure versus 66.1\% for Python-FT, with divergence growing monotonically through the network. Yet \textbf{PyLang-FT and Python-FT remain highly similar to each other} (CKA~0.97--1.00 at all layers): both models develop equivalent internal representations regardless of target language. We note that CKA measures geometric similarity and does not guarantee functional equivalence; however, combined with the behavioral evidence from the LLM judge (\S\ref{sec:judge}), these results provide converging evidence that both models arrive at similar internal problem-solving strategies. The bottleneck is not what the model knows, but what it can express.

\section{Related Work}
\label{sec:related}

\textbf{Code Generation Benchmarks.}
HumanEval~\citep{chen2021evaluating} and MBPP~\citep{austin2021program} are the most widely used benchmarks for evaluating code generation, but they increasingly suffer from data contamination~\citep{zhang2024gsm1k} and pattern-matching exploitation~\citep{gupta2024mmlu}. More challenging benchmarks include LiveCodeBench~\citep{jain2024livecodebench}, SWE-bench~\citep{jimenez2024swebench}, and LiveCodeBench Pro~\citep{zheng2025livecodebenchpro}, where Olympiad medalists annotate problems and find that frontier models still score 0\% on hard problems, succeeding primarily on implementation-heavy tasks rather than those requiring nuanced algorithmic reasoning. MultiPL-E~\citep{cassano2023multipl} broadened evaluation to 18 programming languages, and M2G-Eval~\citep{xu2025m2geval} introduced multi-granularity evaluation across 18 languages, finding strong cross-language correlations that suggest models learn transferable programming concepts. Our work differs from all of these by evaluating the \emph{same} problems across two languages, one known, one unseen to directly isolate what fine-tuning contributes beyond pretraining.

\textbf{Cross-Lingual Code Transfer.} \cite{cassano2023multiplt} use semi-synthetic data to boost low-resource language performance; \cite{hadhoud2026ideafirst} show that generating plans before code improves solve rates but a persistent gap remains. Plan-then-code methods~\citep{wang2024plansearch,jiang2023selfplanning,sun2024codeplan} improve pass rates by up to 25\%. \cite{baltaji2023crosslingual} show cross-lingual transfer outperforms zero-shot across 41 languages; \cite{giagnorio2025nosilverbullet} find fine-tuning helps smaller models but can hurt larger ones. All study \emph{existing} languages that appear in pretraining. Our unseen language has zero pretraining contamination, cleanly isolating the fine-tuning signal.

\textbf{Esoteric Languages as Reasoning Probes.} \cite{sharma2026esolangbench} show models scoring 85--95\% on standard benchmarks drop to 0--11\% on esoteric equivalents. \cite{han2025reward} argue LLM reasoning overfits to training syntax, advocating RL-from-scratch pretraining. We share this motivation but differ by \emph{designing} PyLang as a Python/JavaScript hybrid that is easy to parse but hard to program in, by \emph{fine-tuning} rather than testing zero-shot, and by diagnosing the bottleneck as implementation fidelity rather than reasoning transfer.

\textbf{Interpretability.} \cite{yin2025neuronguided} find lower layers encode language-specific syntax while middle layers capture shared semantics across languages, corroborating our CKA analysis. \cite{basu2026interpretability} demonstrate a knowledge-action gap where internal representations encode knowledge far exceeding output performance, paralleling our finding that PyLang and Python models share nearly identical representations (CKA $>$ 0.97) yet produce substantially different outputs.

For a comprehensive discussion of related work, see Appendix~\ref{appendix:related_work}.

\section{Conclusion}
\label{sec:conclusion}

A central question in LLM code generation is whether the algorithmic reasoning acquired during pretraining can transfer to languages the model has never seen. The research presented in this paper shows that the answer is nuanced: models quickly learn the syntax of a novel language through fine-tuning, but consistently fail to produce semantically correct implementations. We use PyLang, a purpose-built language absent from all pretraining corpora, to study this gap across frontier and open-source models, finding that Python outperforms PyLang by up to 19\% despite identical training data and that no intervention closes the gap. An LLM judge reveals that frontier models select identical algorithms 80\% of the time yet cannot translate it into working PyLang code, confirming that the deficit lies in implementation, not reasoning. Results of this research open the door to more systematic study of cross-language transfer in code generation and to the development of training methods that decouple algorithmic reasoning from language-specific realization.

\section*{Acknowledgements}
The authors thank  Shamik Roy and Yingfan Wang for helpful discussions and feedback on the draft.

\section*{Ethics Statement}
This work introduces a synthetic programming language (PyLang) designed solely for research purposes and does not involve human subjects, personal data, or sensitive information. The benchmark problems are drawn from publicly available competitive programming platforms (Codeforces) and existing academic benchmarks (MBPP). Our LLM judge evaluations use commercially available API services under their standard terms of use. We acknowledge that insights from this work about LLM transfer limitations could inform both beneficial applications (improving code assistants for underserved languages) and potential misuse (identifying weaknesses in code generation systems). We believe the benefits of understanding these limitations outweigh the risks. We plan to release PyLang, the interpreter, and the benchmark to facilitate reproducibility and further research.

\bibliographystyle{colm2026_conference}
\bibliography{references}

\newpage
\appendix
\textbf{\Large Supplement for ``Syntax Without Semantics: Teaching LLMs to Code in an Unseen Language''}

\section{Limitations and Future Work}
PyLang shares surface-level syntactic conventions with C and JavaScript (curly braces, semicolons) but lacks the vast majority of their features, including closures, prototypes, iterators, and standard libraries. Since even this partial familiarity does not close the gap, a language with wholly unfamiliar syntax would likely fare worse. We evaluate only the Qwen 3 family for fine-tuning, though the frontier model results (Sonnet 4, 4.5, GPT 5.4) suggest the phenomenon is not model-specific. Our benchmark draws exclusively from competitive programming and utility tasks, and the gap may manifest differently in other domains such as systems programming or data processing. Additionally, PyLang's interpreter is implemented in Python and inherits Python's runtime characteristics (e.g., arbitrary-precision integers, dynamic typing), so the semantic distance between the two languages is smaller than it would be for a language with fundamentally different execution semantics such as a stack-based or concatenative language.

These limitations point to several directions for future work. First, evaluating on languages with fundamentally different paradigms (e.g., stack-based or functional languages) would test whether the implementation fidelity gap generalizes beyond imperative syntax. Second, testing across additional model families (e.g., Llama, DeepSeek-Coder) would strengthen the generalizability of our findings. Third, extending the benchmark beyond competitive programming to domains such as web development, data pipelines, or embedded systems would reveal whether the gap is uniform or domain-dependent. Finally, developing training methods that explicitly separate algorithmic planning from language-specific code emission, for example through modular architectures or language-conditioned decoding, represents the most promising path toward closing the gap we identify. We release PyLang and the accompanying benchmark as a testbed for progress toward this goal.

\section{Additional Ablation and Error Decomposition Results}
\label{app:sonnet4}

\subsection{Sonnet~4 Prompt Ablation}

Table~\ref{tab:ablation_sonnet4} reports prompt ablation results for Sonnet~4 on PyLang. The same qualitative pattern observed for Sonnet~4.5 (Table~\ref{tab:ablation}) holds: code snippets are the single most important component, and rules alone without snippets actually \emph{hurt} performance (8.2\% vs.\ 12.5\% base). One configuration-specific finding is that the instruction ``use print for output'' is actively harmful, dropping performance from 46.6\% to 22.2\%, likely because it causes the model to overuse print statements in ways that interfere with output formatting. Even in the best configuration, Sonnet~4 reaches only 46.6\% in PyLang versus 85.5\% in Python, a 39\% gap consistent with the frontier model results reported in the main text.

\subsection{Detailed Error Decomposition}

Table~\ref{tab:decomp_detail} provides a fine-grained breakdown of error categories within the PyLang--Python gap across all five models. Several patterns are worth noting. First, chained array indexing (\texttt{arr[i][j]}) is overwhelmingly a Sonnet~4 problem (61 instances vs.\ 21 for Sonnet~4.5 and $\leq$1 for fine-tuned models), confirming that Sonnet~4.5 handles this constraint substantially better, yet its overall gap barely shrinks because other barriers (particularly same-algorithm wrong output) fill the vacuum. Second, same-algorithm wrong output is the single largest language barrier category for every model except Sonnet~4, where chained indexing dominates. Third, fine-tuned models show very few syntax-level barriers ($\leq$6 per model across all syntax categories combined), confirming that fine-tuning effectively teaches PyLang grammar. The dominant failure mode for fine-tuned models is algorithmic divergence (43--57 problems for 8B and 32B), consistent with the judge analysis in \S\ref{sec:judge}: when implementation becomes difficult, fine-tuned models tend to produce a different approach rather than persisting with the correct one.

\begin{table}[h]
\centering
\caption{Sonnet~4 prompt configurations on PyLang (352 problems).
The same qualitative pattern as Sonnet~4.5 holds: code snippets
are essential.}
\label{tab:ablation_sonnet4}
\small
\begin{tabular}{lcc}
\toprule
\textbf{Prompt Configuration} & \textbf{PyLang Pass \%} & \textbf{Syn.\ Err} \\
\midrule
Base (spec only)                       & 12.5 & 600 \\
+ Rules only (no snippets)             &  8.2 & 621 \\
+ Snippets + rules + ``use print''     & 22.2 & 510 \\
+ Snippets + rules (no ``use print'')  & 46.6 & 237 \\
\midrule
\multicolumn{3}{l}{\textit{Python baseline (same model)}} \\
\midrule
Sonnet 4 Python                        & 85.5 & 0 \\
\bottomrule
\end{tabular}
\end{table}

\begin{table*}[t]
\centering
\caption{Detailed error categories within the PyLang--Python gap, by model.
Each row shows the number of gap problems (Python correct, PyLang incorrect)
attributed to a specific failure mode. \textit{Language barriers} involve the
same algorithm; \textit{algorithmic gaps} involve a different algorithm.}
\label{tab:decomp_detail}
\small
\begin{tabular}{llccccc}
\toprule
& \textbf{Error Category} & \textbf{Sonnet 4.5} & \textbf{Sonnet 4} & \textbf{32B} & \textbf{8B} & \textbf{4B} \\
\midrule
\multicolumn{7}{l}{\textit{Language barriers (same algorithm)}} \\
\midrule
& Wrong output: same algo      & 38 & 26 & 22 & 13 & 14 \\
& Syntax: chained indexing      & 21 & 61 &  1 &  0 &  0 \\
& Wrong output: empty output    & 12 &  2 &  0 &  0 &  1 \\
& Runtime: unsupported modulo   &  5 &  3 &  4 &  4 &  9 \\
& Timeout: same algo (slow)     &  4 &  2 &  3 &  0 &  1 \\
& Syntax: unsupported keywords  &  2 & 14 &  1 &  0 &  1 \\
& Syntax: unsupported modulo    &  1 &  3 &  3 &  3 &  4 \\
& Runtime/syntax: other         &  1 &  6 &  2 &  6 & 12 \\
\midrule
& \textbf{Total language barriers} & \textbf{84} & \textbf{117} & \textbf{36} & \textbf{26} & \textbf{42} \\
\midrule
\multicolumn{7}{l}{\textit{Algorithmic gaps (different algorithm)}} \\
\midrule
& Wrong output: diff algo       & 18 & 14 & 41 & 55 & 24 \\
& Timeout: diff algo             &  0 &  3 &  2 &  2 &  1 \\
\midrule
& \textbf{Total algorithmic gaps} & \textbf{18} & \textbf{17} & \textbf{43} & \textbf{57} & \textbf{25} \\
\midrule
& Ambiguous (judge error)       & 10 & 12 & 11 & 15 & 16 \\
\midrule
& \textbf{Total gap}            & \textbf{112} & \textbf{146} & \textbf{90} & \textbf{98} & \textbf{83} \\
\bottomrule
\end{tabular}
\end{table*}



\subsection{The Gap Is Not a Length or Cascading-Error Artifact}\label{app:length}
PyLang solutions are 3--5$\times$ longer than their Python equivalents (Appendix~\ref{app:examples}, Example~3), raising the concern that the gap is a long-text hallucination effect. We stratify by Python reference-solution length as a difficulty proxy (Python length tracks difficulty more cleanly than PyLang length, which conflates difficulty with per-problem verbosity). If longer outputs drove the gap it should grow with length; instead both pass rates fall together and the gap stays roughly flat, and the largest gap (22~pp) appears at the \emph{short} end (Table~\ref{tab:lengthcontrol}), the opposite of a cascading-error prediction. This corroborates the difficulty analysis (largest gap on easy problems, smallest on hard) and the 12--17\% gap on the 238 low-stdlib problems, where PyLang and Python solutions are similar in length.

As an inverse control we ran \emph{Restricted Python}: the base model solves the same problems in Python under PyLang-equivalent restrictions (no imports, no stdlib, no \texttt{for}, no \texttt{int}/\texttt{split}). On the 86 problems where Qwen3-4B complied, Restricted-Python pass rate was ${\sim}$63\%, versus 36.1\% for fine-tuned PyLang. With the standard library equally unavailable in both, writing in the familiar language is still ${\sim}$27~pp more accurate, isolating implementation fidelity as the residual when expressive power is equalized.

\begin{table}[h]
\centering
\caption{PyLang--Python gap by Python reference-solution length (difficulty proxy). The gap does not grow with length.}
\label{tab:lengthcontrol}
\small
\begin{tabular}{lcccc}
\toprule
\textbf{Python len (chars)} & \textbf{N} & \textbf{Python \%} & \textbf{PyLang \%} & \textbf{Gap} \\
\midrule
0--200     & 64 & 65.6 & 60.9 & 4.7 \\
200--500   & 59 & 74.6 & 52.5 & 22.0 \\
500--1000  & 81 & 44.4 & 38.3 & 6.2 \\
1000--2000 & 86 & 27.9 & 23.3 & 4.7 \\
2000--5000 & 59 & 16.9 & 10.2 & 6.8 \\
5000+      &  3 &  0.0 &  0.0 & 0.0 \\
\bottomrule
\end{tabular}
\end{table}

\section{Hyperparameters}
\label{app:hyperparams}

\subsection{Preference Tuning}

\paragraph{Direct Preference Optimization (DPO).}
Preference pairs are constructed by sampling 8 completions per training problem from the Q$\to$C SFT checkpoint at temperature~0.7.
Each completion is evaluated against the test suite, and passing solutions are paired with failing solutions (chosen vs.\ rejected).
Problems where all 8 completions either pass or fail are excluded.
Training proceeds for 4 epochs with $\beta = 0.1$, learning rate $5 \times 10^{-7}$, and full fine-tuning (no LoRA).

\paragraph{Group Relative Policy Optimization (GRPO).}
At each training step, the model generates 8 candidate solutions at temperature~0.7 and receives execution-based rewards.
The reward function is $r = \text{tests\_passed} / \text{tests\_total}$, yielding 1.0 for fully correct solutions, 0.0 for syntax errors, timeouts, or complete failure, and intermediate values for partial correctness.
Training uses 1{,}000 Codeforces problems sampled from the training set, running for 4 epochs with $\beta = 0.1$, learning rate $2 \times 10^{-6}$, and full fine-tuning.

\section{Training Data Generation}
\label{app:datagen}
All 2{,}816 PyLang training and test examples were generated by Claude Opus~4.5 in an agentic loop and kept only if they passed every test case under the actual PyLang interpreter, out of 9{,}269 source Codeforces problems attempted (a 30.4\% yield). The generation prompt supplies the full PyLang interpreter source as the only language specification, the natural-language problem statement, the exact function signature from the first test case, and three few-shot examples auto-selected from earlier verified solutions in the same run, decoded at temperature~0.1, top-$k$~500, with a 4{,}096-token cap. Each candidate is run through the same interpreter used for evaluation with a 45-second per-test limit, and is kept only if every test case passes, since partial passes are discarded. The model may decline a problem with the literal token \texttt{OOS} when it cannot be expressed in PyLang, and such problems are dropped, accounting for about 14\% of pool failures rather than wrong answers. On failure the model gets up to two more attempts, each with a short summary (``Tests failed: $K/N$ passed'') and the interpreter's actual error appended to the prompt.

\section{PyLang Language Reference}
\label{app:pylang}

This appendix provides a complete reference for the PyLang programming language, described in Section~\ref{sec:pylang}.

\subsection{Lexical Structure}

PyLang source code is tokenized into numbers, strings, identifiers, keywords, and operators.

\paragraph{Keywords.} Six reserved words: \texttt{function}, \texttt{return}, \texttt{print}, \texttt{if}, \texttt{else}, \texttt{while}.

\paragraph{Identifiers.} Sequences of alphanumeric characters and underscores, beginning with a letter or underscore.

\paragraph{Literals.}
\begin{itemize}[nosep]
  \item \textbf{Numbers:} Integer and floating-point literals (e.g., \texttt{42}, \texttt{3.14}). Integers are stored as Python \texttt{int} (arbitrary precision); floats as Python \texttt{float}.
  \item \textbf{Strings:} Double-quoted, with escape sequences \texttt{\textbackslash n}, \texttt{\textbackslash t}, \texttt{\textbackslash\textbackslash}, and \texttt{\textbackslash"}.
  \item \textbf{Array literals:} Bracket-delimited, comma-separated expressions (e.g., \texttt{[1, 2, 3]}), stored internally as Python dictionaries mapping integer indices to values.
\end{itemize}

\paragraph{Operators.}
\begin{itemize}[nosep]
  \item \textbf{Arithmetic:} \texttt{+}, \texttt{-}, \texttt{*}, \texttt{/}, \texttt{\%}
  \item \textbf{Comparison:} \texttt{==}, \texttt{!=}, \texttt{<}, \texttt{>}, \texttt{<=}, \texttt{>=}
  \item \textbf{Logical:} \texttt{\&\&}, \texttt{||}
  \item \textbf{Unary:} \texttt{-} (negation, parsed as \texttt{0 - expr})
\end{itemize}

\noindent Operator precedence (lowest to highest): logical $\to$ relational $\to$ additive $\to$ multiplicative $\to$ unary/primary.




\subsection{Semantics}

\paragraph{Variables and scoping.} Variables are dynamically typed and do not require declaration. Assignment creates a variable in the current scope. PyLang uses \textbf{lexical scoping} with a locals stack: variables inside a function are local; variables outside any function are global. Local variables shadow globals. There is no explicit variable declaration syntax, the first assignment creates the variable.

\paragraph{Arrays.} Arrays are implemented as Python dictionaries mapping integer keys to values. This design choice has several consequences: (1) arrays are sparse, assigning \texttt{a[100] = 5} does not allocate indices 0 to 99; (2) \texttt{len()} returns the number of assigned keys, not the maximum index; (3) accessing an uninitialized index returns a large negative sentinel value (\texttt{-sys.maxsize}) rather than raising an error. Arrays are stored in a separate namespace from scalar variables.

\paragraph{Strings.} Strings support concatenation via \texttt{+}, repetition via \texttt{*} (with an integer operand), character-level indexing via \texttt{s[i]}, and length via \texttt{len(s)}. Out-of-bounds indexing returns the empty string. There is no \texttt{substring()}, \texttt{split()}, \texttt{indexOf()}, or any other string method.

\paragraph{Functions.} Functions are defined with the \texttt{function} keyword. Parameters are passed by value for scalars and by reference (shared dictionary) for arrays. Functions may call other user-defined functions and may recurse. The only built-in function is \texttt{len()}, which returns the length of a string or the number of keys in an array.

\paragraph{Control flow.} \texttt{if}/\texttt{else} and \texttt{while} are the only control structures. There is no \texttt{for} loop, \texttt{break}, \texttt{continue}, or \texttt{switch}. All iteration must be expressed with \texttt{while} and explicit counter management. \texttt{return} exits the current function and optionally returns a value.

\paragraph{Type coercion.} PyLang performs no implicit type coercion. Adding a string and an integer raises a \texttt{TypeError}. There is no \texttt{int()}, \texttt{str()}, or \texttt{toString()} function. All type conversion must be performed manually (e.g., digit-by-digit parsing for string-to-integer).

\paragraph{I/O model.} The standard entry point is \texttt{function solve(input)}, where \texttt{input} is the entire stdin content as a single string. Output is produced via \texttt{print()}, which writes its argument followed by a newline. There is no formatted output or input reading beyond the initial \texttt{input} parameter.

\subsection{Operations Requiring Manual Implementation}

Table~\ref{tab:pylang_missing} summarizes the operations that models must implement manually in PyLang, along with the typical code cost.

\begin{table}[h]
\centering
\caption{Common operations that require manual implementation in PyLang. Number of code line estimates are based on typical model-generated solutions.}
\label{tab:pylang_missing}
\small
\begin{tabular}{lll}
\toprule
\textbf{Operation} & \textbf{Python} & \textbf{PyLang (manual)} \\
\midrule
String $\to$ integer & \texttt{int(s)} & 10--15 lines (digit-by-digit) \\
Split by delimiter & \texttt{s.split()} & 10--15 lines (char loop) \\
Sort an array & \texttt{sorted(a)} & 15--25 lines (bubble/insertion) \\
Integer $\to$ string & \texttt{str(n)} & 8--12 lines (modulo loop) \\
Array slicing & \texttt{a[i:j]} & 5--8 lines (while loop + copy) \\
Min/max of array & \texttt{min(a)} & 5--8 lines (linear scan) \\
String contains & \texttt{x in s} & 5--10 lines (nested loop) \\
\bottomrule
\end{tabular}
\end{table}

\subsection{Example Model-Generated Solutions}
\label{app:examples}

Below we show representative solutions generated by our fine-tuned models, illustrating both the verbose input-parsing boilerplate that PyLang requires and the algorithmic core that follows.

\paragraph{Example 1: String validation (Codeforces, passing).} This solution checks whether every consonant (except ``n'') is followed by a vowel. Note the manual vowel checking via repeated \texttt{||} comparisons. PyLang has no \texttt{in} operator or sets.

\begin{lstlisting}[language=pylang, numbers=none]
function solve(input) {
    s = "";
    i = 0;
    while (i < len(input)) {
        if (input[i] != "\n") {
            s = s + input[i];
        }
        i = i + 1;
    }
    i = 0;
    while (i < len(s) - 1) {
        current_char = s[i];
        next_char = s[i + 1];
        if (current_char != "n") {
            is_consonant = 1;
            if (current_char == "a" || current_char == "o"
                || current_char == "u" || current_char == "i"
                || current_char == "e") {
                is_consonant = 0;
            }
            if (is_consonant == 1) {
                is_vowel_next = 0;
                if (next_char == "a" || next_char == "o"
                    || next_char == "u" || next_char == "i"
                    || next_char == "e") {
                    is_vowel_next = 1;
                }
                if (is_vowel_next == 0) {
                    print("NO");
                    return 0;
                }
            }
        }
        i = i + 1;
    }
    print("YES");
}
\end{lstlisting}

\paragraph{Example 2: Check all characters identical (MBPP, passing).} A simpler MBPP example showing basic control flow. The model correctly returns 1 (true) or 0 (false) since PyLang has no boolean type.

\begin{lstlisting}[language=pylang, numbers=none]
function solve(input) {
    if (input == "") {
        return 0;
    }
    first_char = input[0];
    i = 1;
    while (i < len(input)) {
        if (input[i] != first_char) {
            return 0;
        }
        i = i + 1;
    }
    return 1;
}
\end{lstlisting}

\paragraph{Example 3: Input parsing boilerplate.} The most distinctive feature of model-generated PyLang code is the verbose input-parsing prologue. Since PyLang lacks \texttt{int()}, \texttt{split()}, and \texttt{input()}, every Codeforces solution must begin with manual line splitting and digit-by-digit integer conversion. The following pattern appears almost identically at the start of every generated solution:

\begin{lstlisting}[language=pylang, numbers=none]
lines = [];
current_line = "";
i = 0;
line_count = 0;
while (i < len(input)) {
    if (input[i] == "\n") {
        lines[line_count] = current_line;
        line_count = line_count + 1;
        current_line = "";
    } else {
        current_line = current_line + input[i];
    }
    i = i + 1;
}
if (len(current_line) > 0) {
    lines[line_count] = current_line;
}

n = 0;
j = 0;
while (j < len(lines[0])) {
    digit = 0;
    if (lines[0][j] == "0") { digit = 0; }
    if (lines[0][j] == "1") { digit = 1; }
    if (lines[0][j] == "2") { digit = 2; }
    if (lines[0][j] == "3") { digit = 3; }
    if (lines[0][j] == "4") { digit = 4; }
    if (lines[0][j] == "5") { digit = 5; }
    if (lines[0][j] == "6") { digit = 6; }
    if (lines[0][j] == "7") { digit = 7; }
    if (lines[0][j] == "8") { digit = 8; }
    if (lines[0][j] == "9") { digit = 9; }
    n = n * 10 + digit;
    j = j + 1;
}
\end{lstlisting}

\noindent In Python, the equivalent is simply \texttt{n = int(input())}. This 35-line boilerplate, which appears in virtually every Codeforces solution, illustrates why PyLang solutions are 3--5$\times$ longer than their Python equivalents and why each additional line of manual implementation creates an opportunity for the model to introduce a bug.

\section{Comprehensive Related Work}
\label{appendix:related_work}
\paragraph{Code Generation Benchmarks.}
HumanEval~\citep{chen2021evaluating} and MBPP~\citep{austin2021program} are the most widely used benchmarks for evaluating code generation, but they increasingly suffer from data contamination~\citep{zhang2024gsm1k} and pattern-matching exploitation~\cite{gupta2024mmlu}. More challenging benchmarks include LiveCodeBench~\citep{jain2024livecodebench}, SWE-bench~\citep{jimenez2024swebench}, and LiveCodeBench Pro~\citep{zheng2025livecodebenchpro}, where Olympiad medalists annotate problems and find that frontier models still score 0\% on hard problems, succeeding primarily on implementation-heavy tasks rather than those requiring nuanced algorithmic reasoning. MultiPL-E~\citep{cassano2023multipl} broadened evaluation to 18 programming languages, and M2G-Eval~\citep{xu2025m2geval} introduced multi-granularity evaluation across 18 languages, finding strong cross-language correlations that suggest models learn transferable programming concepts. Our work differs from all of these by evaluating the \emph{same} problems across two languages, one known, one unseen, to directly isolate what fine-tuning contributes beyond pretraining.

\paragraph{Cross-Lingual Code Transfer and Low-Resource Languages.}
A growing body of work studies how well code LLMs generalize to languages with limited training data.
~\cite{cassano2023multiplt} propose MultiPL-T, which uses semi-synthetic data translated from Python to boost performance on Julia, Lua, OCaml, R, and Racket.
~\cite{baltaji2023crosslingual} conduct an extensive empirical study of cross-lingual transfer across up to 41 programming languages, demonstrating that transfer significantly outperforms zero-shot learning and that source-target language similarity predicts transfer success.
~\cite{joel2024survey} survey 111 papers on LLM code generation for low-resource and domain-specific languages, categorizing improvement methods and finding that no standard benchmark or approach exists.
~\cite{giagnorio2025nosilverbullet} show there is no silver bullet: fine-tuning helps smaller LLMs on low-resource languages but can actually \emph{hurt} very large models, while in-context learning provides modest but reliable gains.
~\cite{shen2026illagent} propose Inference-time Language Acquisition (ILA-agent), where an LLM masters an unfamiliar language through dynamic interaction with documentation and execution environments.
~\cite{chen2022transferability} study transfer from multilingual to monolingual PLMs for Ruby, finding that strategic language selection reduces fine-tuning time while improving performance.
These works all study \emph{existing} languages that appear, however sparsely, in pretraining corpora.
Our approach creates a \emph{novel} language with zero pretraining contamination, enabling clean isolation of the fine-tuning signal, something impossible with any existing language, however low-resource.

\paragraph{Esoteric and Novel Languages as Reasoning Probes.}
The most directly related line of work uses unconventional programming languages to test whether LLMs reason or memorize.
~\cite{sharma2026esolangbench} introduce EsoLang-Bench, a benchmark using five esoteric languages (Brainfuck, Befunge-98, Whitespace, Unlambda, Shakespeare), finding that models scoring 85--95\% on standard benchmarks drop to 0--11\% on equivalent esoteric tasks, with 0\% accuracy beyond the Easy tier.
~\cite{han2025reward} use esoteric programming languages to demonstrate that LLM reasoning overfits to training data syntax, proposing that the core issue is coupling of knowledge and reasoning in next-token prediction, and advocating RL-from-scratch pretraining.
Singer and ~\cite{singer2025esolangs} provide the most comprehensive academic treatment of esoteric languages, arguing they have pedagogical value by enforcing deep conceptual transfer and exposing raw computation, and noting that LLMs fail at both synthesizing esolang code and designing new esolangs.
Our work shares the motivation of using an unfamiliar language to probe LLM capabilities, but differs in three key ways:
(1)~we create a \emph{purpose-built} novel language rather than reusing existing esolangs, eliminating any residual pretraining contamination;
(2)~we \emph{fine-tune} models rather than testing zero-shot, enabling us to measure what is and is not learnable from limited data;
and (3)~our interpretability analysis reveals that the bottleneck is not reasoning transfer (which succeeds internally) but \emph{implementation fidelity}, a diagnosis distinct from the reasoning-knowledge coupling hypothesis of ~\cite{han2025reward}. ~\cite{budnikov2025synthetic} also study transfer from synthetic languages, though for natural language understanding rather than code, showing that fine-tuning on a well-designed synthetic language can improve English task performance.

\paragraph{Interpretability of Code Language Models.}
Our interpretability analysis builds on methods for understanding how fine-tuning reshapes model representations.
Most directly relevant, ~\cite{yin2025neuronguided} conduct a neuron-level analysis of code LLMs (Llama-3.1-8B, Qwen2.5-Coder-32B) across five programming languages, finding that lower layers encode language-specific syntax while middle layers capture semantic abstractions shared across languages, a finding our CKA and logit lens analyses corroborate at the representation level.
~\cite{phang2021finetuned} use CKA to study fine-tuned transformers on NLU tasks, discovering block-diagonal similarity structures where later layers can be discarded without performance loss.
~\cite{davari2022cka} formally characterize CKA's sensitivity properties, showing it can be manipulated without changing functional behavior and calling for caution when interpreting CKA values, we address this by complementing CKA with SVD rank analysis, logit lens, and activation patching.
~\cite{basu2026interpretability} demonstrate a knowledge-action gap where models encode task-relevant knowledge in internal representations far exceeding their output performance, and mechanistic interpretability methods fail to bridge this gap, closely paralleling our finding that PyLang and Python models share nearly identical internal representations (CKA $>$ 0.97) yet produce substantially different outputs.

\section{Prompt Configurations}
\label{app:prompts}

We report the full prompts used for each ablation configuration evaluated on frontier models. All prompts begin with the problem statement, followed by the components listed below. The base configuration includes only the PyLang interpreter source in Python; subsequent configurations add components incrementally.

\newtcolorbox{promptbox}[1]{
  enhanced,
  breakable,
  colback=blue!3!white,
  colframe=blue!40!black,
  coltitle=white,
  fonttitle=\bfseries,
  title=#1,
  boxrule=0.5pt,
  arc=2pt,
  left=6pt, right=6pt, top=4pt, bottom=4pt,
  fontupper=\ttfamily\small,
}

\begin{promptbox}{Base Prompt (Interpreter Only)}
You are an expert programmer. You are given a programming
language called PyLang, defined by the following interpreter
defined in python:\newline

\{interpreter\_source\}\newline

Write solutions in PyLang. Output ONLY the PyLang code
starting with 'function solve(input) \{'. No explanations.
\end{promptbox}

\begin{promptbox}{Prompt with Interpreter and Rules}
You are an expert programmer. You are given a programming
language called PyLang, defined by the following interpreter
defined in python:\newline

\{interpreter\_source\}\newline

Write solutions in PyLang. Output ONLY the PyLang code
starting with 'function solve(input) \{'. No explanations.
\newline
\newline
RULES:\newline
- Comments are not supported in PyLang. Do not write any comments.\newline
- Double indexing like arr[i][j] is not supported. Use: temp = arr[i]; val = temp[j]
\end{promptbox}

\begin{promptbox}{Prompt with Code Snippets}
You are an expert programmer. You are given a programming
language called PyLang, defined by the following interpreter
defined in python:\newline

\{interpreter\_source\}\newline

The input is always given as a single string. Use these code
snippets as needed based on the problem's Input Format:\newline

If the input is a single string, use it directly:\newline
\quad s = input;\newline

To split input into lines (when input has multiple lines):\newline
\quad lines = [];\newline
\quad current\_line = "";\newline
\quad i = 0;\newline
\quad line\_count = 0;\newline
\quad while (i < len(input)) \{\newline
\quad\quad if (input[i] == "\textbackslash n") \{\newline
\quad\quad\quad lines[line\_count] = current\_line;\newline
\quad\quad\quad line\_count = line\_count + 1;\newline
\quad\quad\quad current\_line = "";\newline
\quad\quad \} else \{\newline
\quad\quad\quad current\_line = current\_line + input[i];\newline
\quad\quad \}\newline
\quad\quad i = i + 1;\newline
\quad \}\newline
\quad if (len(current\_line) > 0) \{\newline
\quad\quad lines[line\_count] = current\_line;\newline
\quad \}\newline

To split a string into space-separated tokens:\newline
\quad tokens = [];\newline
\quad token\_count = 0;\newline
\quad current\_token = "";\newline
\quad k = 0;\newline
\quad while (k < len(line)) \{\newline
\quad\quad if (line[k] == " ") \{\newline
\quad\quad\quad if (len(current\_token) > 0) \{\newline
\quad\quad\quad\quad tokens[token\_count] = current\_token;\newline
\quad\quad\quad\quad token\_count = token\_count + 1;\newline
\quad\quad\quad\quad current\_token = "";\newline
\quad\quad\quad \}\newline
\quad\quad \} else \{\newline
\quad\quad\quad current\_token = current\_token + line[k];\newline
\quad\quad \}\newline
\quad\quad k = k + 1;\newline
\quad \}\newline
\quad if (len(current\_token) > 0) \{\newline
\quad\quad tokens[token\_count] = current\_token;\newline
\quad \}\newline

To convert a string to a number:\newline
\quad s = lines[0];\newline
\quad val = 0;\newline
\quad j = 0;\newline
\quad while (j < len(s)) \{\newline
\quad\quad digit = 0;\newline
\quad\quad if (s[j] == "0") \{ digit = 0; \}\newline
\quad\quad ...\newline
\quad\quad if (s[j] == "9") \{ digit = 9; \}\newline
\quad\quad val = val * 10 + digit;\newline
\quad\quad j = j + 1;\newline
\quad \}\newline

To append to an array, use indexed assignment:\newline
\quad arr[arr\_count] = value;\newline
\quad arr\_count = arr\_count + 1;\newline

Write solutions in PyLang. Output ONLY the PyLang code
starting with 'function solve(input) \{'. No explanations.
\end{promptbox}












\end{document}